\documentclass[conference]{IEEEtran}
\IEEEoverridecommandlockouts
\usepackage{cite}
\usepackage{amsmath,amssymb,amsfonts}
\usepackage{algorithmic}
\usepackage[linesnumbered,ruled]{algorithm2e}
\usepackage{graphicx}
\usepackage{textcomp}
\usepackage{xcolor}
\usepackage{etoolbox,siunitx}
\usepackage{url}
\usepackage{subcaption}
\def\BibTeX{{\rm B\kern-.05em{\sc i\kern-.025em b}\kern-.08em
		T\kern-.1667em\lower.7ex\hbox{E}\kern-.125emX}}
\begin{document}

	\title{A Hybrid GA-PSO Method for Evolving Architecture and Short Connections of Deep Convolutional Neural Networks}

	\author{
		\IEEEauthorblockN{Bin Wang, Yanan Sun, Bing Xue and Mengjie Zhang}
		\IEEEauthorblockA{School of Engineering and Computer Science\\
			Victoria University of Wellington,
			PO Box 600, Wellington 6140, NEW ZEALAND}
		\IEEEauthorblockA{Emails: wangbin@myvuw.ac.nz, \{yanan.sun, bing.xue, mengjie.zhang\}@ecs.vuw.ac.nz}
	}

	\maketitle

	\begin{abstract}
		Image classification is a difficult machine learning task, where Convolutional Neural Networks (CNNs) have been applied for over 20 years in order to solve the problem. In recent years, instead of the traditional way of only connecting the current layer with its next layer, shortcut connections have been proposed to connect the current layer with its forward layers apart from its next layer, which has been proved to be able to facilitate the training process of deep CNNs. However, there are various ways to build the shortcut connections, it is hard to manually design the best shortcut connections when solving a particular problem, especially given the design of the network architecture is already very challenging.
        In this paper, a hybrid evolutionary computation (EC) method is proposed to \textit{automatically} evolve both the architecture of deep CNNs and the shortcut connections. Three major contributions of this work are: Firstly, a new encoding strategy is proposed to encode a CNN, where the architecture and the shortcut connections are encoded separately; Secondly, a hybrid two-level EC method, which combines particle swarm optimisation and genetic algorithms, is developed to search for the optimal CNNs; Lastly, an adjustable learning rate is introduced for the fitness evaluations, which provides a better learning rate for the training process given a fixed number of epochs. The proposed algorithm is evaluated on three widely used benchmark datasets of image classification and compared with 12 peer Non-EC based competitors and one EC based competitor. The experimental results demonstrate that the proposed method outperforms all of the peer competitors in terms of classification accuracy.
	\end{abstract}


\section{Introduction}

Image classification is a difficult machine learning task due to a couple of reasons. Firstly, the dimensionality of the input image is very high. The image is composed of a number of pixels each of which is one dimension. Assuming a grayscale image with size of 256$\times$256 pixels (real-life images are often even larger), the dimensionality is 65, 536, which brings a huge search space for machine learning algorithm.
 Furthermore, the diversity of images in the same class can be large. In order to correctly distinguish the images of various classes, the variability of images in the same class needs to be minimised and the variability of images between different classes has to be maximised. The large diversity of images in the same class makes it extremely difficult to minimise the variability of images in the same class, which therefore causes the complication in image classification.

Deep Convolutional Neural Networks (CNNs) have been the leading approach for solving image classifications tasks since it was introduced around 30 years ago \cite{Backprop_LeCun}. Various CNN methods have been developed, e.g. VGGNet \cite{VGGNet_Simonyan}, Xception \cite{Xception_Chollet} and GoogLeNet \cite{GoogleNet_Szegedy}.
Deep CNNs have achieved better and better accuracy on image classification tasks. However, the architectures of CNNs grow deeper and deeper (i.e. more and more layers), which makes the training of deep CNNs much harder due to the difficulty in the CNNs \textit{architecture design} and \textit{network training}.

Almost all of the state-of-the-art CNNs are with a manually designed architecture, which is very challenging to achieve without expertise both in CNNs and domain knowledge on the target problem. However, most real-world users often do not have such knowledge.
In recent years, evolutionary computation (EC) has shown to be effective in \textit{automatically} searching for the optimal architecture of CNNs \cite{CGP_Miller} \cite{GeneticCNN_Xie} \cite{Yanan2018CNNarchitecture}.

\begin{figure}[t!]
    \centering
    \vspace{-1mm}
    \includegraphics[width=1\linewidth]{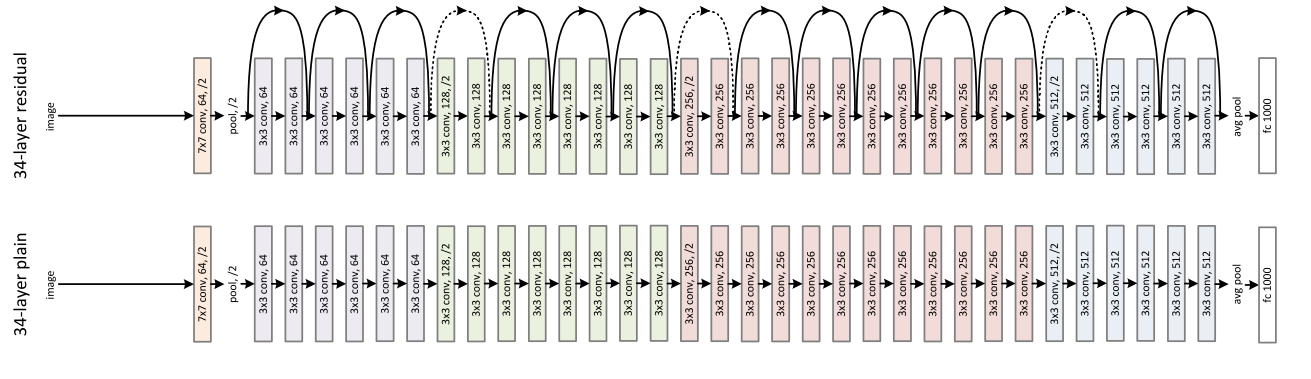}
    \vspace{-6mm}
    \caption{ResNet architecture  (image taken from \cite{Resnet_he})}
    \label{fig:resnet_architecture}
\end{figure}

\begin{figure}[t!]
    \vspace{-2mm}
    \centering
    \includegraphics[width=1\linewidth]{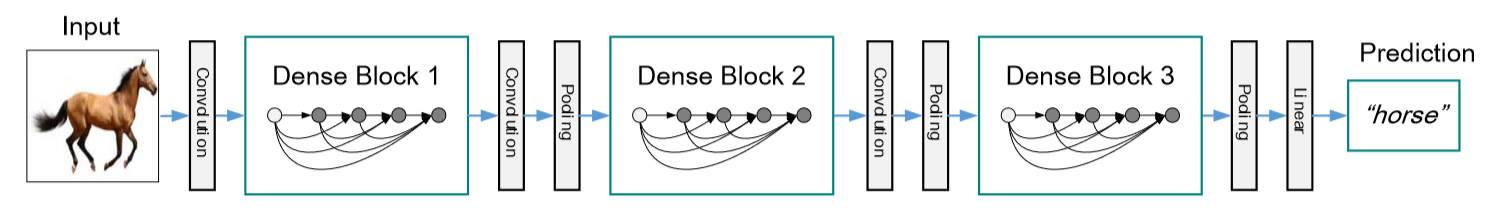}
    \vspace{-3mm}
    \caption{DenseNet architecture (image taken from \cite{Densenet_huang})}
    \label{fig:densenet_architecture}
     \vspace{-1mm}
\end{figure}

Back-propagation with gradient descent optimisation is the most commonly used method for training CNNs, but the vanishing gradients problem often occurs when training a deep CNN \cite{Bengio_GradientDescent} \cite{Srivastava_TrainingVeryDeepNN}.
Recently, \textit{shortcut connections} have been introduced and shown to be effective in dealing with this problem \cite{orhan2018skip}.
Shortcut connections add extra connections between the current layer and the forward layers. Typical examples are ResNet \cite{Resnet_he} as shown in Fig. \ref{fig:resnet_architecture} and the densely-connected shortcuts in DenseNet \cite{Densenet_huang} as illustrated in Fig. \ref{fig:densenet_architecture}. As can be seen from Fig. \ref{fig:resnet_architecture}, in ResNet, along with the direct forward connections between the current layer and the next layer, there are also jump connections, which connect the current layer to the layer after the next layer. DenseNet divides the CNN architecture into a number of blocks. Each layer can be connected to all of the forward layers of the same block, which is called densely-connected structure. Such shortcut connections have been heavily investigated in recent years with different variants \cite{orhan2018skip} \cite{Jin_SkipConnection}. However, such shortcut connections are manually designed and there still are a large number of open questions. For example, although the operations after shortcut connections are addition in ResNet and concatenation in DenseNet, it is unclear whether the shortcut connections in ResNet with the concatenation mechanism is better than DenseNet. Without rich expertise, it is still challenging to design the best shortcut connections to effectively and efficiently address a given problem. Therefore, it is needed to develop an approach to automatically searching for the shortcut connections.





\textbf{Goals:}
we aim to develop a novel EC based approach that can automatically find the optimal CNN architecture and decide whether there should be shortcut connection(s) between one layer and its forward layer(s).
A two-level encoding strategy is proposed, which is then used by a hybrid EC method of a genetic algorithm (GA) and particle swarm optimisation (PSO) to evolve both the network architecture and shortcut connections.
Since both the architecture and the shortcut connections are dynamically decided during the evolutionary process without any human interference, the proposed method is named \textit{DynamicNet}. The proposed method will be examined and compared with one EC based method and 12 state-of-the-art non-EC based methods on three of the widely-used datasets having different levels of difficulties. The specific objectives and contributions are:

\begin{itemize}
	\item Design a new encoding strategy that includes both the CNN architecture and the shortcut connections. Since the CNN architecture is decisive to the classification accuracy and the shortcut connections impact how well the CNN can be trained, a two-level encoding is proposed with the first level representing the CNN architecture and the second level representing the shortcut connections. These two levels are encoded as a vector with decimal values and a vector of binary values, respectively;
	\item Develop a hybrid algorithm that can work with the two-level encoding. A variable-length PSO algorithm is proposed to evolve the CNN architectures due to PSO's promising performance on continuous optimisation while GA is used to evolve the shortcut connections since it works well on optimisation tasks with binary values;
	\item Propose a new fitness evaluation method to improve the effectiveness and efficiency of the encoded CNN. Classification accuracy is used as the fitness value of the proposed method. Each evaluation requires to train the encoded CNN, which is an expensive process. Motivated by previous work \cite{YananEvoCNN}, a small number of training epochs is used to speed up the training. Furthermore, an automation method is developed to search for the best learning rate among a sequence of learning rates to improve the classification accuracy.
\end{itemize}


\section{Background}\label{sec:Background}

\subsection{ResNet}\label{S:background_ResNet}

As shown in Fig. \ref{fig:resnet_architecture}, the architecture at the bottom represents a plain CNN architecture called VGG nets \cite{VGG_Simonyan}, which mostly contains convolutional layers with 3$\times$3 filters; while by inserting shortcut connections, the plain architecture is turned into the recently proposed ResNet shown as the architecture at the top of the figure. The output is calculated based on Equation (\ref{eq:ResNet_ouput}), where $x$ is the input, $\mathcal{F}(x, {W_{i}})$ represents the output of the convolutional layer with the weights $W_{i}$, and $W_{s}$ can be a constant of \textbf{1} if the dimension of the input is identical to that of the output of the convolutional layer; otherwise it will be a linear projection of the input in order to match the dimension of the output of the convolutional layer.

\begin{equation}\label{eq:ResNet_ouput}
y = \mathcal{F}(x, {W_{i}}) + W_{s}x
\end{equation}

\subsection{DenseNet}\label{S:background_DenseNet}

\begin{figure}[ht!]
	\centering
	\includegraphics[width=1\linewidth]{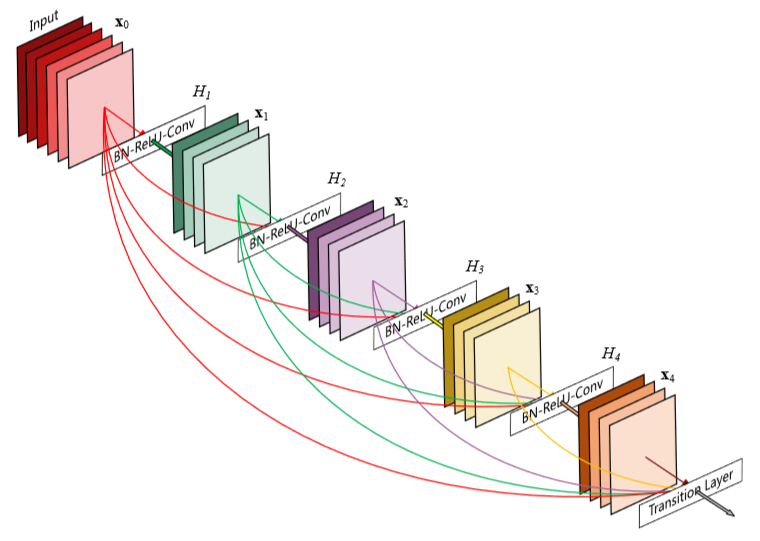}
	\caption{A five layer dense block (image taken from \cite{Densenet_huang})}
	\label{fig:dense_block}
\end{figure}

DenseNet is a newly proposed CNN architecture in image classification tasks. As shown in Fig. \ref{fig:densenet_architecture}, a DenseNet is composed of several dense blocks, and the convolutional layer and the pooling layer between the dense blocks which are referred to as the transition layer. Fig. \ref{fig:dense_block} illustrates a dense block, which contains five layers, and a transition layer at the end. To be more specific with the dense block, suppose a single image $x_{0}$ is passed to a dense block, which is composed of $L$ layers. Each of the $L$ layers implements a non-linear transformation $H_{l}(\cdot)$, and the output of the $l^{th}$ layer is denoted as $x_{l}$. As the output of the $l^{th}$ layer receives all of the feature maps of all preceding layers, the output $x_{l}$ can be calculated according to Formula (\ref{eq:DenseNet_ouput}), where $[x_{0}, x_{1}, ..., x_{l-1}]$ refers to the concatenation of the feature maps obtained from layer 0, 1, ..., $l-1$, and $H_{l}$ represents a composite function of three consecutive operations, which are batch normalization (BN) \cite{BN_Ioffe}, a rectified linear unit (ReLU) \cite{ReLU_Glorot} and $3\times3$ convolution (Conv).

\begin{equation}\label{eq:DenseNet_ouput}
	x_{l} = H_{l}([x_{0}, x_{1}, ..., x_{l-1}])
\end{equation}

\subsection{GAs and PSO}\label{S:background_GA_PSO}
\paragraph{GAs} As an EC approach, GAs are inspired by the process of natural selection. The bio-inspired operators, such as mutation, crossover and selection, are utilised to evolve the population in order to obtain a high-quality solution \cite{GAIntro_mitchell}. The procedure of GA is composed of five parts: initialisation, selection, fitness evaluation, mutation, and crossover. At the stage of initialisation, a population of random vectors with a fixed dimension is generated; Next, the selection is performed by using a selection algorithm to select the individuals into a mating pool; After that, mutation is performed by selecting one individual from the mating pool and the value of each dimension is randomly chosen to be changed to evolve a new individual;
Crossover is performed by selecting two individuals in the mating pool and combining a part of one individual's vector with that of the other. By iterating the fitness evaluation, selection, mutation, and crossover, the new population can be filled with new individuals with hopefully better solutions.
The whole process terminates when the stopping criteria are met, and the best individual of all generations is reported as the evolved solution.

\paragraph{PSO}
As one of the EC approaches, PSO is motivated by the social behaviour of fish schooling or bird flocking \cite{PSOIntro_Kennedy} \cite{PSOIntro_Eberhart_Y}. In PSO, there is a population consisting of a number of candidate solutions also called particles, and each particle has a position and a velocity. The representation of the position is $\mathbf{x_{i}} = (x_{i1}, x_{i2},...x_{id})$, where $\mathbf{x_{i}}$ is a vector of a fixed dimension representing the position of the $i\mathit{th}$ particle in the population and $x_{id}$ means the $d\mathit{th}$ dimension of the $i\mathit{th}$ particle's position. $\mathbf{v_{i}} = (v_{i1}, v_{i2},...v_{id})$ illustrates the velocity of a particle, where $\mathbf{v_{i}}$ is a fix-length vector expressing the velocity of the $i\mathit{th}$ particle and $v_{id}$ means the $d\mathit{th}$ dimension of the $i\mathit{th}$ particle's velocity. The way that PSO solves the optimisation problems is to keep moving the particle to a new position in the search space until the stopping criteria are met. The position of the particle is updated according to the update equation which incorporates two equations - the velocity update equation \ref{eq:PSO_UpdateV} and the position update Equation (\ref{eq:PSO_UpdateX}). In Formula (\ref{eq:PSO_UpdateV}), $v_{id}(t+1)$ indicates the updated $d\mathit{th}$ dimension of the $i\mathit{th}$ particle's velocity, $r_{1}$ and $r_{2}$ carry random numbers between 0 and 1, $w, c_{1}$ and $c_{2}$ are PSO parameters that are used to fine-tune the performance of PSO, and $P_{id}$ and $P_{gd}$ bear the $d\mathit{th}$ dimension of the local best and the global best, respectively. After updating the velocity of the particle, the new position can be achieved by applying Formula (\ref{eq:PSO_UpdateX}).

\begin{equation}\label{eq:PSO_UpdateV}
\resizebox{.9\hsize}{!}{
	$v_{id}(t+1) = w * v_{id}(t) + c_{1} * r_{1} * (P_{id} - x_{id}(t)) + c_{2} * r_{2} * (P_{gd} - x_{id}(t))$
}
\end{equation}
\begin{equation}\label{eq:PSO_UpdateX}
x_{id}(t+1) = x_{id}(t) + v_{id}(t+1)
\end{equation}

\subsection{Related Work}\label{S:background_related_work}
With regard to the automatic design of artificial neural networks, two research areas haven been surged in recent years, which are neuroevolution (NE) and neural architecture search (NAS) \cite{Pablo_MemeticEvolution} \cite{NAS_Zoph}. The traditional NE only evolves the connection weights given a fixed topology. NeuroEvolution of Augmenting Topologies (NEAT) \cite{NEAT_Stanley} was proposed to evolve neural network topologies along with weights, which outperformed the best fixed-topology method on a challenging benchmark reinforcement learning task. On top of NEAT, a Hypercube-based NeuroEvoltuion of Augmenting Topologies (HyperNEAT) was developed to open up a new class of complex high-dimensional tasks to neuroevolution. However, neuroevolution is hard to be applied on evolving deep CNNs because the search space of deep CNNs, which combines the topologies and the connection weights, is too large, which makes EC algorithms hard to perform well. As a result, NAS \cite{NAS_Zoph} was proposed to automatically search for the optimal topology of CNNs, but not to evolve the connection weights. Instead, the connection weights were trained by backpropagation. However, the process of training a deep CNN is slow, which results in a very slow NAS process. In the proposed NAS with reinforcement learning \cite{NAS_Zoph}, the experiments ran on 500 GPUs for 28 days to obtain the final result, and the improved NAS method called NASNet \cite{NASNet_Zoph} has dramatically reduced the running time, but the experiments still took 4 days on 500 GPUs.

Since the computational cost of NAS is extremely high, and most of the research institutes do not have such powerful GPU resource of 500 GPUs, in the past years, interested researchers have proposed quite a lot of NAS algorithms to evolve CNN architectures with a much lower computational cost, e.g. in \cite{PSOCNN_Wang} \cite{Masanori_GPCNN} \cite{DECNN_Wang} \cite{Pablo_MemeticEvolution}. However, in the proposed algorithms with much lower computational cost, the classification accuracy is comprised a little bit. For example, the proposed method in \cite{DECNN_Wang} does not achieve an ideal result on one of the benchmark datasets called CONVEX.

Although there has been a large amount of work done in automatically designing neural networks without any human interference, there is often a trade-off between the efficiency and effectiveness. Along with recent CNNs becoming deeper and deeper, automatically evolving CNNs becomes much harder, especially with limited hardware resources.


\section{The Proposed Method}\label{C:dynamicnet_algorithms}
The proposed method is described in this section, which uses a hybrid EC method of GA and PSO (HGAPSO). The overall structure of the proposed method is described, followed by the details of the algorithms. The new CNN architecture named DynamicNet\footnote{DynamicNet is implemented as one of the models in the Python library:\url{https://pypi.org/project/convtt/}}, which will be evolved by the HGAPSO algorithm, is introduced. Then the encoding strategy to represent DynamicNet is designed as the rest of the algorithm is dependent on them. Based on that, the overview of the algorithm, the first-level PSO evolution and the second-level GA evolution are described; Lastly, the fitness evaluation is defined.

\subsection{Overall Structure of the System}\label{S:algorithms_hgapso_System}
\begin{figure}[!t]
    \centering
    \includegraphics[width=0.9\linewidth]{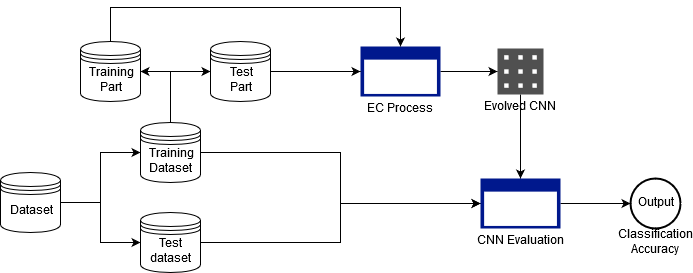}
    \caption{The flowchart of the experimental process}
    \label{fig:expriment_workflow}
    \vspace{-2mm}
\end{figure}

Fig. \ref{fig:expriment_workflow} shows the overall structure of the system (this structure is actually used by all of the experiments in this paper). The dataset is split into a training set and a test set, and the training set is further divided into a training part and a test part. The training part and the test part are passed to the EC process, which is the HGAPSO algorithm. During the fitness evaluations, the training part is used to train the neural network, and the test part is used to obtain the test accuracy of the trained neural network, which is used as the fitness value. EC produces the evolved CNN architecture, which is the best individual. Lastly, in the CNN evaluation procedure, the produced CNN architecture is trained on the whole training set, and the test accuracy of the trained CNN model is obtained, which is the final output of the system.

\subsection{DynamicNet - The Evolved CNN Architecture}
By comparing the figures of ResNet and DenseNet, it can be observed that in ResNet, each layer has at most two connections from previous layers. However, in DenseNet, the connections of each layer coming from previous layers are the number of previous layers due to the densely-connected structure. Therefore, the number of input feature maps is the sum of the numbers of feature maps of all previous layers, which results in the exploding growth of the number of feature maps, particularly for the layers near the output layer. The solution introduced in DenseNet is to divide the whole CNN into multiple blocks called Dense Block as shown in Fig. \ref{fig:dense_block}. Each block is followed by a transition layer, which comprises a convolutional layer and a pooling layer, to reduce the number of feature maps to half the number of input feature maps. The hyperparameters of the convolutional layer are fixed, which are 3 as the filter size, 1 as the stride size, and half the number of input feature maps as the number of feature maps; The pooling layer also has fixed hyperparameters, which are 2$\times$2 as the kernel size and 2 as the stride size. As the proposed DynamicNet may be densely-connected, it might have the same exploding growth issue of the number of feature maps. Therefore, DynamicNet adopts the block mechanism of DenseNet.

Inside each block, there are a number of convolutional layers with a fixed filter size of 3$\times$3 and a fixed stride size of 1. After each layer, the total number of input feature maps grows by the number of feature maps of the convolutional layer, which is called the \textit{growth rate} of the block. In DenseNet, \textit{the number of blocks}, \textit{the number of convolutional layers} and \textit the {growth rate} are manually designed, which requires good domain knowledge and a lot of manual trials to find a good architecture. In the proposed HGAPSO algorithm, these three hyperparameters will be also automatically designed.

\subsection{HGAPSO Encoding Strategy}\label{S:algorithm_hgapso_encoding}

DynamicNet is comprised of a number of blocks which are connected by transition layers, and the shortcut connections are built between layers inside the block. Based on the construction pattern of the network, the hyperparameters of the architecture can be split into the architecture and the shortcut connections. Regarding the architecture of the network, there are various hyperparameters including the number of blocks, the number of convolutional layers in each block and the growth rate of the convolutional layer in the block, which need to be evolved. In addition to the densely-connected structure in DenseNet, different topologies of shortcut connections, i.e. the different combination of partial shortcut connections in each block, will be explored by the proposed HGAPSO method in order to keep the meaningful features and remove the unmeaningful features learned by previous layers.

\begin{figure}[ht!]
	\centering
	\begin{subfigure}[b]{.5\textwidth}
		\centering
		\includegraphics[width=.9\textwidth]{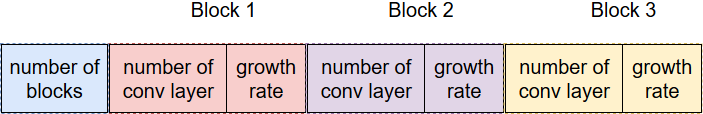}
		\caption{HGAPSO first-level encoding}
		\label{fig:hgapso_encoding_1}
	\end{subfigure}
	\begin{subfigure}[b]{.3\textwidth}
		\centering
		\includegraphics[width=\textwidth]{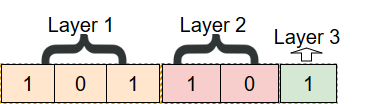}
		\caption{HGAPSO second-level encoding}
		\label{fig:hgapso_encoding_2}
	\end{subfigure}
	\caption{HGAPSO encoding}
	\vspace{-2mm}
\end{figure}

Based on the analysis of the architecture and hyperparameters, the encoding process can be divided into two steps. The first step is to encode the hyperparameters of the CNN architecture. Each of the hyperparameters is a dimension of the architecture encoding, which is shown in Fig. \ref{fig:hgapso_encoding_1}. The first dimension is the number of blocks, and the two hyperparameters of each block, the number of convolutional layers and the growth rate, as two dimensions are appended to the vector. The first step of the encoding is named the first-level encoding, which will be used by the first-level evolution. Based on the results of the first-level encoding, the shortcut connections can be encoded into a binary vector at the second-level encoding. An example of one block with 5 layers is illustrated in Fig. \ref{fig:hgapso_encoding_2}. Each of the dimensions represents a shortcut connection between two layers that are not next to each other, and the two layers next to each other are always connected. Taking the first layer as an example, the three binary digits - [101] represents the shortcut connections between the first layer to the third, to the fourth, and to the fifth layer, respectively, where 1 means the connection exists and 0 means there is no connection. A number of similar binary vectors shown in Fig. \ref{fig:hgapso_encoding_2} constitute the whole vector that represents the shortcut connections of the whole block.

\subsection{HGAPSO Search }\label{S:algorithms_hgapso_overview}
\subsubsection{Overview}
Based on the two-level encoding strategy, the algorithm is composed of two levels of evolution, which are described in Algorithm \ref{alg:hgapso_framework}. The first-level evolution is designed to evolve the architecture of the CNNs encoded by the first-level encoding, and the second-level evolution is performed to search for the best combination of shortcut connections. There are a couple of reasons to separate the architecture evolution from the evolution of the shortcut-connection combination. First of all, since the architecture and the shortcut connections play different roles in the performance of CNNs, which are that the architecture including the depth and the width of the CNNs represents the capacity of network and the shortcut connections are to facilitate the training process of the network, the training process is only comparable when the architecture is fixed, which inspires the idea of splitting the evolution to two levels. Secondly, if the hyperparameters of both the architecture and the shortcut connections are combined into one encoded vector, it will bring some uncertainties to the search space, which, therefore, may deteriorate the complex search space by introducing more disturbance to the search space.

It is arguable that the computational cost of the two-level evolution may be high, but the two-level encoding strategy divides the complex search space into two smaller search spaces and reduces the disturbance in the search space, so the two-level evolution we believe will not perform worse than searching for the optima in a much more complex search space. Other than that, as the second-level evolution of searching for the best combination of shortcut connections only depends on the specific architecture evolved in the first-level evolution, the second-level evolution can be done in parallel for each of the individual of the first-level evolution, which can dramatically speed up the process if sufficient hardware is available.

\begin{algorithm}[ht!]
	\caption{Framework of HGAPSO}
	\label{alg:hgapso_framework}
	\begin{algorithmic}
		\renewcommand{\algorithmicrequire}{\textbf{Input:}}
		\renewcommand{\algorithmicensure}{\textbf{Output:}}
		\STATE $P \leftarrow$ Initialize the population with first-level encoding elaborated in Section \ref{S:algorithm_hgapso_encoding};
		\STATE $P_{best} \leftarrow$ Empty PSO Personal Best;
		\STATE $G_{best} \leftarrow$ Empty PSO global best;
		\WHILE{first-level termination criterion is not satisfied}
		\STATE $P \leftarrow$ Update the population with first-level PSO evolution described in Section \ref{S:algorithms_hgapso_pso_update};
		\FOR{particle $ind$ in population $P$}
		\STATE $P\_sub \leftarrow$ Initialize the population with second-level encoding based on the value of $ind$ illustrated in Section \ref{S:algorithm_hgapso_encoding};
		\WHILE{second-level termination criterion is not satisfied}
		\STATE $P\_sub \leftarrow$ Update the population with second-level GA evolution described in Section \ref{S:algorithms_hgapso_ga_update};
		\STATE evaluate the fitness value of each individual;
		\STATE $P\_sub_{best} \leftarrow$ retrieve the best individual in $P\_sub$;
		\ENDWHILE
		\STATE Update $P_{best}$ if $P\_sub_{best}$ is better than $P_{best}$;
		\ENDFOR
		\STATE $G_{best} \leftarrow$ retrieve the best individual in $P$;
		\ENDWHILE
	\end{algorithmic}
\end{algorithm}

\subsubsection{HGAPSO First-level PSO evolution}\label{S:algorithms_hgapso_pso_update}
Algorithm \ref{alg:hgapso_pso_update} shows the pseudo code of the PSO evolutionary process.
Based on the encoded vector from the first-level encoding, the value of each dimension is a decimal value, and PSO has been proved to be effective and efficient in solving optimisation problems with decimal values, so PSO is chosen to perform the first-level search. However, the dimensionality of the encoded vector is not fixed, so an adapted variable-length PSO is proposed to solve this variable-length problem. Since the size of the input feature maps to each block is different and the specific block is trained and designed to learn meaningful features given the size of the input feature maps, when applying EC operators on two individuals, it is important to find the matched blocks which have the same size of input feature maps and apply the operators on the matched blocks. To be specific with the PSO evolution in HGAPSO, the length of the particle may be different from the length of the personal best and global best, so based on the blocks of the individual, the corresponding blocks in the personal best and the global best need to be matched by selecting the blocks with the same size of the output feature and the PSO algorithm is only applied on the matched blocks.

The first dimension of the vector represents the number of blocks. When the number of blocks changes, the depth of the CNN architectures changes, which achieves the ability to evolve the depth of the CNN architecture and keeping the diversity of the PSO population. However, the change of the number of blocks incurs a dramatic change to the CNN architecture, and if it changes too often, each CNN architecture evolution might be too short to achieve good performance, so it is better to leave the evolution some time to optimise other hyperparameters given the specific number of blocks. In order to keep the diversity of the number of blocks and reduce the disturbance caused by frequently changing the number of blocks, the rate of changing the number of blocks in the vector is introduced, which is a real value between [0, 1]. Therefore, the preference for diversity or stability depending on specific tasks can be controlled by tweaking the rate of changing the number of blocks.

When the number of blocks is changed, some blocks need to be randomly cut or randomly generated in order to meet the requirement of the number of blocks in the first dimension. For example, suppose the number of blocks is increased from 3 to 4, the hyperparameters of the fourth block need to be randomly generated based on the first-level encoding strategy, which then are appended to the vector of 3 blocks; On the other way around, assuming the number of blocks is decreased from 4 to 3, the last block is removed. In the proposed HGAPSO method, whenever removing a block(s), it always starts from the last layer because it does not affect the feature map sizes of the other blocks.

\begin{algorithm}[ht!]
	\caption{HGAPSO first-level PSO evolution}
	\label{alg:hgapso_pso_update}
	\begin{algorithmic}
		\renewcommand{\algorithmicrequire}{\textbf{Input:}}
		\renewcommand{\algorithmicensure}{\textbf{Output:}}
		\REQUIRE The current particle $ind$, the personal best $P_{best}$, the global best $G_{best}$, the rate of changing the number of blocks $r_{cb}$;
		\STATE $rnd \leftarrow$ Generate a random number from a uniform distribution;
		\STATE Find the matched blocks of the particle $ind$ by comparing the feature map size;
		\STATE Update the velocity and position of the matched blocks of the particle $ind$ according to Equation \ref{eq:PSO_UpdateV} and \ref{eq:PSO_UpdateX};
		\IF{$rnd < r_{cb}$}
		\STATE Update the velocity and position of the dimension of number of blocks of the particle $ind$ according to Equation \ref{eq:PSO_UpdateV} and \ref{eq:PSO_UpdateX};
		\STATE Randomly cut or generate the blocks to the value of the number of blocks.
		\ENDIF
	\end{algorithmic}
\end{algorithm}

\subsubsection{HGAPSO Second-level GA evolution}\label{S:algorithms_hgapso_ga_update}
According to the second-level encoding depicted in Section \ref{S:algorithm_hgapso_encoding}, once the CNN architecture is obtained from the first-level evolution, the dimensionality of the second-level encoding will be fixed, so the encoded vector can be represented by a fixed-length binary vector. Since GAs are good at optimising binary problems, a GA is chosen as the algorithm to perform the second-level evolution.

\subsection{HGAPSO Fitness Evaluations}\label{S:algorithms_hgapso_fitness}
It can be observed from Algorithm \ref{alg:hgapso_framework} that fitness evaluation only takes place inside the second-level GA evolution, and the fitness of the best GA individual is used as the fitness of its corresponding particle of first-level PSO evolution. Backpropagation with Adam Optimiser \cite{Adam_Kingma} is used to train the network for a number of epochs on part of the training data. The accuracy of the trained CNN on the test part of the training data as the fitness value of the individual.

There are two hyperparameters for the fitness evaluations, which are the number of epochs and the initial learning rate of Adam Optimiser. In the experiment, 5 epochs are used by considering the hardware available and a fairly-short experimental time. After the number of epochs is chosen, DenseNet is used as a baseline to determine an initial learning rate for optimising a CNN with the given depth and width, i.e. after the architecture of the CNN determined, the network with fully-connected blocks as shown in Fig. \ref{fig:dense_block} are used to find a best initial learning rate among 0.9, 0.1 and 0.01.

In order to speed up the evolution process, a part of the training dataset is used for the second-level evolution because the second-level evolution consumes the most computation. While for the first-level evolution, as the computational cost is not that high, and in order to achieve a more stable performance given the architecture of a CNN, the full training dataset is used.


\section{Experiment Design}\label{C:dynamicnet_experiments}

\subsection{Benchmark Datasets and State-of-the-art Competitors}

Due to our limited hardware resource, the DECNN method proposed in \cite{DECNN_Wang}, which only requires a few days running of the experiment on a single GPU, is chosen as the peer EC competitor instead of the method proposed in \cite{NAS_Zoph}, which takes 28 days on 500 GPUs to obtain the final result. The state-of-the-art machine learning algorithms used to compare with DECNN are also used as the peer Non-EC competitors. As DECNN did not perform well on CONVEX benchmark dataset \cite{Benchmarks_Larochelle}, CONVEX dataset is selected as one of the benchmark datasets to see if the proposed HGAPSO algorithm able to achieve better performance. Apart from the CONVEX dataset, the MB and MDRBI datasets \cite{Benchmarks_Larochelle} are also used as benchmark datasets to evaluate the proposed algorithm across different complexities, as MB is the simplest dataset among the MB variants, and MDRBI is the most complicated variant of the MB datasets. On MB, the images represent the handwritten digits from 0 to 9, and there are 12,000 training images and 5,000 test images; MDRBI contains the same amount of training and test images, but some noises are added to the original MB dataset. The CONVEX dataset contains images with the shape of convex or non-convex, which are divided into the training dataset of 8,000 images and the test dataset of 5,000 images. Since EC methods are stochastic, the experiment will be run 30 times and statistical tests will be performed when comparing the proposed algorithm with its peer competitors.

As it would be more convincing to evaluate the proposed HGAPSO algorithm on larger datasets such as CIFAR-10, but the computational cost is too high, e.g. one run of HGAPSO on CIFAR-10 takes more than a week. Therefore, the experiment on CIFAR-10 will not be run for 30 times due to the very high computational cost, our limited GPU resource and the time constraint. Instead, only one run of the experiment will be performed in order to obtain an initial result, which gives suggestions on whether it is worth continuing the experiments for 30 runs in the future when more GPU resources are ready.

\subsection{Parameter settings}\label{S:dynamicnet_ec_parameters}

All of the parameters are configured according to the conventions in the communities of PSO \cite{PSOSettings_Van} and GAs \cite{GAParameter_Digalakis} along with taking into account the computational cost and the complexity of the search space. The values of the parameters of the proposed algorithm are listed in Table \ref{table:hpsoga_ParameterList}.

\begin{table}[!ht]
	\renewcommand{\arraystretch}{1.3}
	\caption{Parameter Settings}
	\label{table:hpsoga_ParameterList}
	\centering
	\scriptsize
	\begin{subtable}{.48\linewidth}
		\begin{tabular}{|p{.48\linewidth}|c|}
    \hline
            Parameter & Value\\
            \hline
            \multicolumn{2}{|l|}{HGAPSO parameters} \\
            \hline
            the range of \# of layers in each block & [4, 8]\\
            \hline
            the range of growth rate in each block & [8, 32]\\
            \hline
            population size & 20\\
            \hline
            generation & 10\\
			\hline
		\end{tabular}
	\end{subtable}%
	\begin{subtable}{.48\linewidth}
		\centering
		\begin{tabular}{|p{.48\linewidth}|c|}
    \hline
            Parameter & Value\\
            \hline
            \multicolumn{2}{|l|}{PSO} \\
            \hline
            $c_1$, $c_2$  & 1.49618\\
            \hline
           $w$   & 0.7298\\
            \hline
            \multicolumn{2}{|l|}{GA} \\
			\hline
			mutation rate & 0.01\\
			\hline
			cross over rate & 0.9\\
			\hline
			elitism rate & 0.1\\
			\hline
		\end{tabular}
	\end{subtable}
\end{table}



\section{Results and Discussions}\label{C:dynamicnet_results}
When comparing the proposed HGAPSO method (stochastic) with the state-of-the-art methods (deterministic), One Sample T-Test is applied to test whether the results of HGAPSO is statistically significantly better, due to that ; when the comparison of error rates between HGAPSO and the proposed DECNN is performed, Two Sample T-test is utilised. Table \ref{table:DynamicNet_VS_SOA} shows the comparison results between HGAPSO and the state-of-the-art algorithms; Table \ref{table:DynamicNet_VS_DECNN} compares HGAPSO with DECNN.

\subsection{HGAPSO vs. State-of-the-Art methods}

The experimental results and the comparison between HGAPSO and the state-of-the-art methods are shown in Table \ref{table:DynamicNet_VS_SOA}. In order to clearly show the comparison results, the terms (+) and (-) are provided to indicate the result of HGAPSO is significantly better or worse than the best result obtained by the corresponding peer competitor; The term (=) shows that the mean error rate of HGAPSO is similar to that of the competitor, i.e. no significant difference from the statistical point of view. The term (--) means there are no available results reported from the provider or cannot be counted.

It can be observed that the proposed HGAPSO method achieves a significant improvement in terms of the error rates shown in Table \ref{table:DynamicNet_VS_SOA}. HGAPSO significantly outperforms the other peer competitors across all the three benchmark datasets. To be specific, it further reduces the error rate over the best competitor by 5\%, 1\% and 10\% on the CONVEX, MB and MDRBI datasets, respectively.

\begin{table}[ht]
	\caption{Classification errors of HGAPSO and Competitors}
	\label{table:DynamicNet_VS_SOA}
	\centering
	\scriptsize
	\begin{tabular}{|m{1.8cm}|S[table-format=2] c|S[table-format=2] c|S[table-format=2] c|}
		\hline
		Method & & CONVEX & & MB & & MDRBI \\
		\hline
		CAE-2  & & -- & 2.48 & (+) & 45.23 & (+) \\
		\hline
		TIRBM  & & -- &  & -- & 35.5 & (+) \\
		\hline
		PGBM+DN-1  & & -- &  & -- & 36.76 & \\
		\hline
		ScatNet-2  & 6.5 & (+) & 1.27 & (+) & 50.48 & (+) \\
		\hline
		RandNet-2  & 5.45 & (+) & 1.25 & (+) & 43.69 & (+) \\
		\hline
		PCANet-2 (softmax)  & 4.19 & (+) & 1.4 & (+) & 35.86 & (+) \\
		\hline
		LDANet-2  & 7.22 & (+) & 1.05 & (+) & 38.54 & (+) \\
		\hline
		SVM+RBF  & 19.13 & (+) & 30.03 & (+) & 55.18 & (+) \\
		\hline
		SVM+Poly  & 19.82 & (+) & 3.69 & (+) & 54.41 & (+) \\
		\hline
		NNet & 32.25 & (+) & 4.69 & (+) & 62.16 & (+) \\
		\hline
		SAA-3 & 18.41 & (+) & 3.46 & (+) & 51.93 & (+) \\
		\hline
		DBN-3 & 18.63 & (+) & 3.11 & (+) & 47.39 & (+) \\
		\hline
		\hline
		HGAPSO(best) & 1.0322090792838878 & & 0.7444852941176521 & & 10.529491687979544 & \\
		\hline
		HGAPSO(mean) & 1.2402379430032862 & & 0.8384700687340176 & & 12.230678548593351 & \\
		\hline
		HGAPSO(standard deviation) & 0.10306204025011284 & & 0.06890141495082609 & & 0.8587440506882343 & \\
		\hline
	\end{tabular}
\end{table}

\subsection{HGAPSO vs. DECNN}

In Table \ref{table:DynamicNet_VS_DECNN}, it can be observed that by comparing the results between HGAPSO and DECNN, both the mean error rate and the standard deviation of HGAPSO are smaller than that of DECNN, and from the statistical point of view, HGAPSO has a significant improvement in terms of the classification accuracy.

\begin{table}[!t]
	\scriptsize
	\caption{Classification rates of HGAPSO and DECNN}
	\label{table:DynamicNet_VS_DECNN}
	\centering
	\begin{tabular}{|m{1.8cm}|S[table-format=1.0] c|S[table-format=1.0] c|S[table-format=1.0] c|}
		\hline
		& & CONVEX & & MB & & MDRBI\\
		\hline
		HGAPSO(mean) & 1.2402379430032862 & & 0.8384700687340176 & & 12.230678548593351 & \\
		\hline
		HGAPSO(standard deviation) & 0.10306204025011284 & & 0.06890141495082609 & & 0.8587440506882343 & \\
		\hline
		DECNN(mean) & 11.1921056069825 &  & 1.45690014362335 & & 37.55115771 & \\
		\hline
		DECNN(standard deviation) & 1.94301999129248 & & 0.113365088 & & 2.446911498 & \\
		\hline
		P-value & \textbf{0.0001} & & \textbf{0.0001} & & \textbf{0.0001} & \\
		\hline
	\end{tabular}
\end{table}

\subsection{Evolved CNN Architecture}

After investigating the evolved CNN architectures, it is found that HGAPSO demonstrates its capability of evolving both the architecture of CNNs and the shortcut connections between layers. By looking into the evolved CNN architectures, it can be observed that not only the CNN architectures with various number of layers but also different topologies of shortcut connections are evolved. For example, one evolved CNN architecture has 3 blocks. In the first block, there are 4 convolutional layers, and [0, 0, 0, 0, 1], [0, 1, 0, 1], [0, 0, 1], [0, 0] and [1] represent the connections from the input, the first layer, the second layer, the third layer to the following layers, where 1 indicates the connection exists, and 0 means no connection; The second block is composed of 8 layers with the growth rate of 34, and the corresponding connections are [1, 0, 1, 0, 1, 0, 1, 0], [0, 1, 1, 1, 1, 0, 1], [1, 1, 1, 1, 1, 0], [1, 1, 1, 0, 1], [1, 0, 0, 0], [0, 0, 0], [1, 1] and [0]; In the third block, there are 5 layers with the corresponding connections of [0, 0, 1, 1, 0], [0, 0, 0, 0], [1, 0, 0], [0, 1] and [0], and the growth rate is 39.

\subsection{One-run Result on CIFAR-10 dataset}

As mentioned earlier, the computational cost of testing HGAPSO is extremely high. For one run of the experiment using one GPU card, it takes more than a week to evolve the CNN architecture, and it took almost 12 hours to train the evolved CNN architecture. The classification accuracy of the specific run is 95.63\%, which ranks the second among the state-of-the-art deep neural networks ranging from 75.86\% to 96.53\% that are collected by the rodrigob website \footnote{http://rodrigob.github.io/are\_we\_there\_yet/build/classification\_datasets\_results.html\#43494641522d3130}; However, all of the state-of-the-art deep neural networks require very highly specialised domain knowledge and tremendous experiments to manually fine-tune the performance, while HGAPSO has the ability of automatically evolving the CNN architecture without any human interference, which is considered as the biggest advantage.


\section{Conclusions}\label{C:hpsoga_con}
This paper developed an EC based method for automatically evolving both the architecture of CNNs and shortcut connections, without human intervention or domain knowledge in either CNNs or the target problem. The proposed method outperforms both the EC competitor and the Non-EC competitors on commonly used benchmark datasets. The first reason is that by evolving shortcut connections, the feature maps learned in previous layers can be reused in further layers, which amplifies the leverage of useful knowledge; Secondly, the shortcut connections make the training of very deep neural networks more effectively by passing the gradients through shortcut connections, which has been proven by DenseNet \cite{Densenet_huang}. Furthermore, the classification accuracy of HGAPSO on CIFAR-10 is promising as it is very competitive with the state-of-the-art deep neural networks. In addition, the most advantage of HGAPSO is that it does not require any human efforts to design the architecture of CNNs, which is usually required for the peer state-of-the-art competitors.

In regard to the future work, there are two aspects came up from the experiments and learnt experience of this paper. Firstly, due to the hardware limitation, the proposed algorithm has been tested on relatively small datasets. Even though an initial result of running HGAPSO on CIFAR-10 is obtained, the statistical analysis based on the results from 30 runs needs to be applied in order to make a stronger claim of the proposed method. It would be more convincing if the algorithms could be tested on other larger datasets such as ImageNet dataset. Secondly, as there are more and more new CNN architectures proposed with better performance, it would be helpful to investigate more recent CNN architectures, based on which EC methods can be applied to automatically evolve more advanced CNN architectures.

\bibliography{dynamicnet}

\begin{thebibliography}{10}
\providecommand{\url}[1]{#1}
\csname url@samestyle\endcsname
\providecommand{\newblock}{\relax}
\providecommand{\bibinfo}[2]{#2}
\providecommand{\BIBentrySTDinterwordspacing}{\spaceskip=0pt\relax}
\providecommand{\BIBentryALTinterwordstretchfactor}{4}
\providecommand{\BIBentryALTinterwordspacing}{\spaceskip=\fontdimen2\font plus
\BIBentryALTinterwordstretchfactor\fontdimen3\font minus
  \fontdimen4\font\relax}
\providecommand{\BIBforeignlanguage}[2]{{%
\expandafter\ifx\csname l@#1\endcsname\relax
\typeout{** WARNING: IEEEtran.bst: No hyphenation pattern has been}%
\typeout{** loaded for the language `#1'. Using the pattern for}%
\typeout{** the default language instead.}%
\else
\language=\csname l@#1\endcsname
\fi
#2}}
\providecommand{\BIBdecl}{\relax}
\BIBdecl

\bibitem{Backprop_LeCun}
Y.~LeCun, B.~Boser, J.~S. Denker, D.~Henderson, R.~E. Howard, W.~Hubbard, and
  L.~D. Jackel, ``Backpropagation applied to handwritten zip code
  recognition,'' \emph{Neural computation}, vol.~1, no.~4, pp. 541--551, 1989.

\bibitem{LeNetHandWritten_LeCun}
------, ``Backpropagation applied to handwritten zip code recognition,''
  \emph{Neural Computation}, vol.~1, no.~4, pp. 541--551, Dec 1989.

\bibitem{LeNetDocument_LeCun}
Y.~Lecun, L.~Bottou, Y.~Bengio, and P.~Haffner, ``Gradient-based learning
  applied to document recognition,'' \emph{Proceedings of the IEEE}, vol.~86,
  no.~11, pp. 2278--2324, Nov 1998.

\bibitem{ImageNet_Alex}
A.~Krizhevsky, I.~Sutskever, and G.~E. Hinton, ``Imagenet classification with
  deep convolutional neural networks,'' \emph{Communications of the ACM},
  vol.~60, no.~6, p. 84–90, 2017.

\bibitem{VGGNet_Simonyan}
\BIBentryALTinterwordspacing
Simonyan, Karen, Zisserman, and Andrew, ``Very deep convolutional networks for
  large-scale image recognition,'' Apr 2015. [Online]. Available:
  \url{https://arxiv.org/abs/1409.1556}
\BIBentrySTDinterwordspacing

\bibitem{Squeezenet_Iandola}
F.~N. Iandola, S.~Han, M.~W. Moskewicz, K.~Ashraf, W.~J. Dally, and K.~Keutzer,
  ``Squeezenet: Alexnet-level accuracy with 50x fewer parameters and< 0.5 mb
  model size,'' \emph{arXiv preprint arXiv:1602.07360}, 2016.

\bibitem{Inception_Szegedy}
C.~Szegedy, V.~Vanhoucke, S.~Ioffe, J.~Shlens, and Z.~Wojna, ``Rethinking the
  inception architecture for computer vision,'' in \emph{Proceedings of the
  IEEE conference on computer vision and pattern recognition}, 2016, pp.
  2818--2826.

\bibitem{Xception_Chollet}
F.~Chollet, ``Xception: Deep learning with depthwise separable convolutions,''
  \emph{arXiv preprint}, pp. 1610--02\,357, 2017.

\bibitem{GoogleNet_Szegedy}
C.~Szegedy, W.~Liu, Y.~Jia, P.~Sermanet, S.~Reed, D.~Anguelov, D.~Erhan,
  V.~Vanhoucke, and A.~Rabinovich, ``Going deeper with convolutions,'' in
  \emph{2015 IEEE Conference on Computer Vision and Pattern Recognition
  (CVPR)}, June 2015, pp. 1--9.

\bibitem{Resnet_he}
\BIBentryALTinterwordspacing
K.~He, X.~Zhang, S.~Ren, and J.~Sun, ``Deep residual learning for image
  recognition,'' \emph{CoRR}, vol. abs/1512.03385, 2015. [Online]. Available:
  \url{http://arxiv.org/abs/1512.03385}
\BIBentrySTDinterwordspacing

\bibitem{Densenet_huang}
\BIBentryALTinterwordspacing
G.~Huang, Z.~Liu, and K.~Q. Weinberger, ``Densely connected convolutional
  networks,'' \emph{CoRR}, vol. abs/1608.06993, 2016. [Online]. Available:
  \url{http://arxiv.org/abs/1608.06993}
\BIBentrySTDinterwordspacing

\bibitem{BN_Ioffe}
S.~Ioffe and C.~Szegedy, ``Batch normalization: Accelerating deep network
  training by reducing internal covariate shift,'' \emph{arXiv preprint
  arXiv:1502.03167}, 2015.

\bibitem{ReLU_Glorot}
X.~Glorot, A.~Bordes, and Y.~Bengio, ``Deep sparse rectifier neural networks,''
  in \emph{Proceedings of the fourteenth international conference on artificial
  intelligence and statistics}, 2011, pp. 315--323.

\bibitem{GAIntro_mitchell}
M.~MITCHELL, \emph{An introduction to genetic algorithms}.\hskip 1em plus 0.5em
  minus 0.4em\relax MIT Press, 1996.

\bibitem{PSOIntro_Kennedy}
J.~Kennedy and R.~Eberhart, ``Particle swarm optimization,'' in \emph{Neural
  Networks, 1995. Proceedings., IEEE International Conference on}, vol.~4, Nov
  1995, pp. 1942--1948 vol.4.

\bibitem{PSOIntro_Eberhart}
R.~Eberhart and J.~Kennedy, ``A new optimizer using particle swarm theory,'' in
  \emph{Micro Machine and Human Science, 1995. MHS '95., Proceedings of the
  Sixth International Symposium on}, Oct 1995, pp. 39--43.

\bibitem{NEAT_Stanley}
K.~O. Stanley and R.~Miikkulainen, ``Evolving neural networks through
  augmenting topologies,'' \emph{Evolutionary computation}, vol.~10, no.~2, pp.
  99--127, 2002.

\bibitem{NAS_Zoph}
B.~Zoph and Q.~V. Le, ``Neural architecture search with reinforcement
  learning,'' \emph{arXiv preprint arXiv:1611.01578}, 2016.

\bibitem{NASNet_Zoph}
B.~Zoph, V.~Vasudevan, J.~Shlens, and Q.~V. Le, ``Learning transferable
  architectures for scalable image recognition,'' \emph{arXiv preprint
  arXiv:1707.07012}, vol.~2, no.~6, 2017.

\bibitem{PSOCNN_Wang}
B.~Wang, Y.~Sun, B.~Xue, and M.~Zhang, ``Evolving deep convolutional neural
  networks by variable-length particle swarm optimization for image
  classification,'' in \emph{2018 IEEE Congress on Evolutionary Computation
  (CEC)}, July 2018, pp. 1--8.

\bibitem{DECNN_Wang}
------, ``A hybrid differential evolution approach to designing deep
  convolutional neural networks for image classification,'' in
  \emph{Australasian Joint Conference on Artificial Intelligence}.\hskip 1em
  plus 0.5em minus 0.4em\relax Springer, 2018, pp. 237--250.

\bibitem{Adam_Kingma}
D.~P. Kingma and J.~Ba, ``Adam: A method for stochastic optimization,''
  \emph{arXiv preprint arXiv:1412.6980}, 2014.

\bibitem{Benchmarks_Larochelle}
H.~Larochelle, D.~Erhan, A.~Courville, J.~Bergstra, and Y.~Bengio, ``An
  empirical evaluation of deep architectures on problems with many factors of
  variation,'' in \emph{Proceedings of the 24th international conference on
  Machine learning}.\hskip 1em plus 0.5em minus 0.4em\relax ACM, 2007, pp.
  473--480.

\bibitem{PSOSettings_Van}
F.~Vandenbergh and A.~Engelbrecht, ``A study of particle swarm optimization
  particle trajectories,'' \emph{Information Sciences}, vol. 176, no.~8, p.
  937–971, 2006.

\bibitem{GAParameter_Digalakis}
J.~Digalakis and K.~Margaritis, ``An experimental study of benchmarking
  functions for genetic algorithms,'' \emph{Proceedings of 2000 IEEE
  International Conference on Systems, Man and Cybernetics.}, 2000.

\end{thebibliography}


\begin{thebibliography}{10}
\providecommand{\url}[1]{#1}
\csname url@samestyle\endcsname
\providecommand{\newblock}{\relax}
\providecommand{\bibinfo}[2]{#2}
\providecommand{\BIBentrySTDinterwordspacing}{\spaceskip=0pt\relax}
\providecommand{\BIBentryALTinterwordstretchfactor}{4}
\providecommand{\BIBentryALTinterwordspacing}{\spaceskip=\fontdimen2\font plus
\BIBentryALTinterwordstretchfactor\fontdimen3\font minus
  \fontdimen4\font\relax}
\providecommand{\BIBforeignlanguage}[2]{{%
\expandafter\ifx\csname l@#1\endcsname\relax
\typeout{** WARNING: IEEEtran.bst: No hyphenation pattern has been}%
\typeout{** loaded for the language `#1'. Using the pattern for}%
\typeout{** the default language instead.}%
\else
\language=\csname l@#1\endcsname
\fi
#2}}
\providecommand{\BIBdecl}{\relax}
\BIBdecl

\bibitem{Backprop_LeCun}
Y.~LeCun, B.~Boser, J.~S. Denker, D.~Henderson, R.~E. Howard, W.~Hubbard, and
  L.~D. Jackel, ``Backpropagation applied to handwritten zip code
  recognition,'' \emph{Neural computation}, vol.~1, no.~4, pp. 541--551, 1989.

\bibitem{VGGNet_Simonyan}
\BIBentryALTinterwordspacing
Simonyan, Karen, Zisserman, and Andrew, ``Very deep convolutional networks for
  large-scale image recognition,'' Apr 2015. [Online]. Available:
  \url{https://arxiv.org/abs/1409.1556}
\BIBentrySTDinterwordspacing

\bibitem{Xception_Chollet}
F.~Chollet, ``Xception: Deep learning with depthwise separable convolutions,''
  \emph{arXiv preprint}, pp. 1610--02\,357, 2017.

\bibitem{GoogleNet_Szegedy}
C.~Szegedy, W.~Liu, Y.~Jia, P.~Sermanet, S.~Reed, D.~Anguelov, D.~Erhan,
  V.~Vanhoucke, and A.~Rabinovich, ``Going deeper with convolutions,'' in
  \emph{2015 IEEE Conference on Computer Vision and Pattern Recognition
  (CVPR)}, June 2015, pp. 1--9.

\bibitem{CGP_Miller}
\BIBentryALTinterwordspacing
J.~Miller and A.~Turner, ``Cartesian genetic programming,'' in
  \emph{Proceedings of the Companion Publication of the 2015 Annual Conference
  on Genetic and Evolutionary Computation}, ser. GECCO Companion '15.\hskip 1em
  plus 0.5em minus 0.4em\relax New York, NY, USA: ACM, 2015, pp. 179--198.
  [Online]. Available: \url{http://doi.acm.org/10.1145/2739482.2756571}
\BIBentrySTDinterwordspacing

\bibitem{GeneticCNN_Xie}
L.~Xie and A.~Yuille, ``Genetic cnn,'' in \emph{2017 IEEE International
  Conference on Computer Vision (ICCV)}, Oct 2017, pp. 1388--1397.

\bibitem{Yanan2018CNNarchitecture}
\BIBentryALTinterwordspacing
Y.~Sun, B.~Xue, M.~Zhang, and G.~G. Yen, ``Automatically designing {CNN}
  architectures using genetic algorithm for image classification,''
  \emph{CoRR}, vol. abs/1808.03818, 2018. [Online]. Available:
  \url{http://arxiv.org/abs/1808.03818}
\BIBentrySTDinterwordspacing

\bibitem{Bengio_GradientDescent}
Y.~Bengio, P.~Simard, and P.~Frasconi, ``Learning long-term dependencies with
  gradient descent is difficult,'' \emph{IEEE transactions on neural networks},
  vol.~5, no.~2, pp. 157--166, 1994.

\bibitem{Srivastava_TrainingVeryDeepNN}
\BIBentryALTinterwordspacing
R.~K. Srivastava, K.~Greff, and J.~Schmidhuber, ``Training very deep
  networks,'' \emph{CoRR}, vol. abs/1507.06228, 2015. [Online]. Available:
  \url{http://arxiv.org/abs/1507.06228}
\BIBentrySTDinterwordspacing

\bibitem{orhan2018skip}
\BIBentryALTinterwordspacing
E.~Orhan and X.~Pitkow, ``Skip connections eliminate singularities,'' in
  \emph{International Conference on Learning Representations}, 2018. [Online].
  Available: \url{https://openreview.net/forum?id=HkwBEMWCZ}
\BIBentrySTDinterwordspacing

\bibitem{Resnet_he}
\BIBentryALTinterwordspacing
K.~He, X.~Zhang, S.~Ren, and J.~Sun, ``Deep residual learning for image
  recognition,'' \emph{CoRR}, vol. abs/1512.03385, 2015. [Online]. Available:
  \url{http://arxiv.org/abs/1512.03385}
\BIBentrySTDinterwordspacing

\bibitem{Densenet_huang}
\BIBentryALTinterwordspacing
G.~Huang, Z.~Liu, and K.~Q. Weinberger, ``Densely connected convolutional
  networks,'' \emph{CoRR}, vol. abs/1608.06993, 2016. [Online]. Available:
  \url{http://arxiv.org/abs/1608.06993}
\BIBentrySTDinterwordspacing

\bibitem{Jin_SkipConnection}
J.~Yamanaka, S.~Kuwashima, and T.~Kurita, ``Fast and accurate image super
  resolution by deep cnn with skip connection and network in network,'' in
  \emph{Neural Information Processing}, D.~Liu, S.~Xie, Y.~Li, D.~Zhao, and
  E.-S.~M. El-Alfy, Eds.\hskip 1em plus 0.5em minus 0.4em\relax Cham: Springer
  International Publishing, 2017, pp. 217--225.

\bibitem{YananEvoCNN}
\BIBentryALTinterwordspacing
Y.~Sun, B.~Xue, and M.~Zhang, ``Evolving deep convolutional neural networks for
  image classification,'' \emph{CoRR}, vol. abs/1710.10741, 2017. [Online].
  Available: \url{http://arxiv.org/abs/1710.10741}
\BIBentrySTDinterwordspacing

\bibitem{VGG_Simonyan}
\BIBentryALTinterwordspacing
K.~Simonyan and A.~Zisserman, ``Very deep convolutional networks for
  large-scale image recognition,'' \emph{CoRR}, vol. abs/1409.1556, 2014.
  [Online]. Available: \url{http://arxiv.org/abs/1409.1556}
\BIBentrySTDinterwordspacing

\bibitem{BN_Ioffe}
S.~Ioffe and C.~Szegedy, ``Batch normalization: Accelerating deep network
  training by reducing internal covariate shift,'' \emph{arXiv preprint
  arXiv:1502.03167}, 2015.

\bibitem{ReLU_Glorot}
X.~Glorot, A.~Bordes, and Y.~Bengio, ``Deep sparse rectifier neural networks,''
  in \emph{Proceedings of the fourteenth international conference on artificial
  intelligence and statistics}, 2011, pp. 315--323.

\bibitem{GAIntro_mitchell}
M.~MITCHELL, \emph{An introduction to genetic algorithms}.\hskip 1em plus 0.5em
  minus 0.4em\relax MIT Press, 1996.

\bibitem{PSOIntro_Kennedy}
J.~Kennedy and R.~Eberhart, ``Particle swarm optimization,'' in \emph{Neural
  Networks, 1995. Proceedings., IEEE International Conference on}, vol.~4, Nov
  1995, pp. 1942--1948 vol.4.

\bibitem{PSOIntro_Eberhart_Y}
Eberhart and Y.~Shi, ``Particle swarm optimization: developments, applications
  and resources,'' in \emph{Proceedings of the 2001 Congress on Evolutionary
  Computation (IEEE Cat. No.01TH8546)}, vol.~1, May 2001, pp. 81--86 vol. 1.

\bibitem{Pablo_MemeticEvolution}
\BIBentryALTinterwordspacing
P.~R. Lorenzo and J.~Nalepa, ``Memetic evolution of deep neural networks,'' in
  \emph{Proceedings of the Genetic and Evolutionary Computation Conference},
  ser. GECCO '18.\hskip 1em plus 0.5em minus 0.4em\relax New York, NY, USA:
  ACM, 2018, pp. 505--512. [Online]. Available:
  \url{http://doi.acm.org/10.1145/3205455.3205631}
\BIBentrySTDinterwordspacing

\bibitem{NAS_Zoph}
B.~Zoph and Q.~V. Le, ``Neural architecture search with reinforcement
  learning,'' \emph{arXiv preprint arXiv:1611.01578}, 2016.

\bibitem{NEAT_Stanley}
K.~O. Stanley and R.~Miikkulainen, ``Evolving neural networks through
  augmenting topologies,'' \emph{Evolutionary computation}, vol.~10, no.~2, pp.
  99--127, 2002.

\bibitem{NASNet_Zoph}
B.~Zoph, V.~Vasudevan, J.~Shlens, and Q.~V. Le, ``Learning transferable
  architectures for scalable image recognition,'' \emph{arXiv preprint
  arXiv:1707.07012}, vol.~2, no.~6, 2017.

\bibitem{PSOCNN_Wang}
B.~Wang, Y.~Sun, B.~Xue, and M.~Zhang, ``Evolving deep convolutional neural
  networks by variable-length particle swarm optimization for image
  classification,'' in \emph{2018 IEEE Congress on Evolutionary Computation
  (CEC)}, July 2018, pp. 1--8.

\bibitem{Masanori_GPCNN}
M.~Suganuma, S.~Shirakawa, and T.~Nagao, ``A genetic programming approach to
  designing convolutional neural network architectures,'' in \emph{Proceedings
  of the Genetic and Evolutionary Computation Conference}.\hskip 1em plus 0.5em
  minus 0.4em\relax ACM, 2017, pp. 497--504.

\bibitem{DECNN_Wang}
B.~Wang, Y.~Sun, B.~Xue, and M.~Zhang, ``A hybrid differential evolution
  approach to designing deep convolutional neural networks for image
  classification,'' in \emph{Australasian Joint Conference on Artificial
  Intelligence}.\hskip 1em plus 0.5em minus 0.4em\relax Springer, 2018, pp.
  237--250.

\bibitem{Adam_Kingma}
D.~P. Kingma and J.~Ba, ``Adam: A method for stochastic optimization,''
  \emph{arXiv preprint arXiv:1412.6980}, 2014.

\bibitem{Benchmarks_Larochelle}
H.~Larochelle, D.~Erhan, A.~Courville, J.~Bergstra, and Y.~Bengio, ``An
  empirical evaluation of deep architectures on problems with many factors of
  variation,'' in \emph{Proceedings of the 24th international conference on
  Machine learning}.\hskip 1em plus 0.5em minus 0.4em\relax ACM, 2007, pp.
  473--480.

\bibitem{PSOSettings_Van}
F.~Vandenbergh and A.~Engelbrecht, ``A study of particle swarm optimization
  particle trajectories,'' \emph{Information Sciences}, vol. 176, no.~8, p.
  937–971, 2006.

\bibitem{GAParameter_Digalakis}
J.~Digalakis and K.~Margaritis, ``An experimental study of benchmarking
  functions for genetic algorithms,'' \emph{Proceedings of 2000 IEEE
  International Conference on Systems, Man and Cybernetics.}, 2000.

\end{thebibliography}
\bibliographystyle{IEEEtran}

\end{document}